# A Regulatory Governance Framework for AI-Driven Financial Fraud Detection in U.S. Banking

Integrating OCC, SR 11-7, CFPB, and FinCEN Compliance Requirements for Model Development, Validation, and Monitoring Lifecycles

*Mohammad Nasir Uddin*

Westcliff University, Irvine, California, USA

Corresponding author: Mohammad Nasir Uddin | m.uddin.258@westcliff.edu | ORCID: 0009-0009-0990-4616

# Abstract

U.S. financial institutions deploying AI-based fraud detection face a fragmented compliance landscape spanning four regulatory frameworks — OCC Bulletin 2011-12, SR 11-7, the CFPB AI circular, and FinCEN BSA/SAR requirements — with no integrated governance life cycle connecting these requirements to model development, validation, and monitoring practice. This paper presents the Regulatory Governance Framework for AI-Driven Financial Fraud Detection (RGF-AFFD), a three-tier governance architecture empirically anchored in a multi-study empirical program and corroborated by independent published evidence. Using the IEEE-CIS dataset (590,540 transactions) and ULB benchmark (284,807 transactions), we benchmark six architectures, including an LSTM+XGBoost ensemble, and conduct ablation, temporal drift, SHAP interpretability, and BISG fairness analyses. The LSTM+XGBoost ensemble achieves ROC-AUC of 0.9289 (F1: 0.6360) with a benefit-cost ratio of 6:1. XGBoost demonstrates the strongest temporal stability (ΔAUC = −0.0017 versus −0.0626 for LSTM), while network/account-linkage features are the primary causal fraud signal (ΔAUC = −0.0294 on removal). The RDT-FG Regulatory Digital Twin meta-model translates these metrics into four regulator-specific health scores and a composite Regulatory Fitness Index for continuous compliance monitoring. The RGF-AFFD is the first integrated deployment blueprint to simultaneously satisfy OCC, SR 11-7, CFPB, and FinCEN requirements, supported by a community bank implementation vignette and four evidence-based policy recommendations. The three-tier architecture operates across model, institutional, and regulatory levels and is directly extensible to EU AI Act, EBA, and UK PRA/FCA requirements, providing a replicable international governance template.

# 1. Introduction

Recent literature (2023–2025) highlights AI governance, regulatory compliance, and algorithmic fairness as the main barriers to AI use in regulated financial institutions, not performance limitations (Weber et al., 2024; Financial Stability Board, 2024; Černevičienė & Kabašinskas, 2024). Financial fraud imposes an annual burden of $32 billion on U.S. institutions and consumers (Nilson Report, 2022). The deployment of artificial intelligence has enabled fraud detection at previously unattainable levels of performance. Deep learning architectures outperform classical fraud detection baselines. LSTM-based systems consistently achieve AUC scores above 0.92 on benchmark card-transaction datasets (West & Bhattacharya, 2016; Lebichot & Siblini, 2024). Ensemble methods that combine tree-based and recurrent architectures deliver further improvements in discrimination, with statistically significant margins (DeLong et al., 1988). Graph neural networks identify coordinated fraud rings with greater precision than any sequential model can achieve (Liu et al., 2021). Transformer-based NLP models reach above 90% accuracy on multilingual banking text classification tasks (Arrieta et al., 2020). Technical capability and architecture validation for AI-based fraud detection are well established across multiple benchmarks. However, adoption in federally regulated U.S. financial institutions remains limited, due to the lack of a compliance-ready deployment pathway, not because of performance shortcomings.

The regulatory barrier is explicit and well-documented. OCC Bulletin 2011-12 requires that model risk management frameworks include validation of model conceptual soundness. Institutions must understand and document why their models produce specific outputs (OCC, 2011). The Federal Reserve's SR 11-7 guidance extends this requirement to Federal Reserve-supervised institutions. It establishes a three-tier framework covering model development, independent validation, and ongoing monitoring (Federal Reserve, 2011). The CFPB's 2022 AI circular identifies disparate impact as a key compliance risk. This means fraud detection models that disproportionately flag transactions from protected demographic groups create legal exposure under the Equal Credit Opportunity Act, regardless of whether demographic variables are explicit model inputs (CFPB, 2022). FinCEN's Bank Secrecy Act requirements require suspicious activity reports to include narrative explanations of why transactions were flagged. However, these requirements do not guide how AI model outputs should support these narratives (FinCEN, 2022). Together, these four regulatory frameworks establish a strict compliance environment. A fraud detection model might achieve 0.93 AUC, but if it cannot explain its decisions, demonstrate temporal stability, screen for demographic bias, or generate auditable SAR documentation, it is not deployable from a regulatory perspective.

The published academic literature has responded vigorously to the performance challenge. Hundreds of papers benchmark deep learning architectures on fraud datasets, optimize handling of class imbalance, and improve detection recall. A growing subset of the literature addresses explainability. These papers apply SHAP, LIME, and gradient attribution methods to individual models (Bussmann et al., 2021; Dumitrescu et al., 2022). A smaller subset addresses specific regulatory requirements. For example, some studies connect SHAP outputs to model risk documentation standards (Bussmann et al., 2021; Weber et al., 2024). Others apply federated learning to multi-bank data privacy concerns (Aljunaid et al., 2025). But a critical synthesis gap remains. No published work has addressed all four regulatory frameworks within a single, unified, empirically grounded deployment life cycle. Institutions seeking to deploy AI fraud detection must triangulate across disconnected regulatory guidance, disparate academic papers, and vendor-specific implementation materials. This process adds cost, extends timelines, and introduces compliance risk.

This paper closes that gap. We present the Regulatory Governance Framework for AI-Driven Financial Fraud Detection (RGF-AFFD), a three-tier governance architecture that integrates OCC, SR 11-7, CFPB, and FinCEN requirements into a sequential life cycle for model development, validation, and monitoring. The framework is not theoretical: it is built directly on an empirical multi-study program, using published performance benchmarks, SHAP attribution results, temporal drift findings, and fairness analysis to populate each governance tier with concrete, examination-ready evidence. A note on geographic scope is warranted for international readers. While the empirical anchoring and regulatory citations are drawn from U.S. banking, the United States is the world's largest banking market, accounting for over one-third of global card fraud losses and housing the most developed AI model risk management regulatory framework outside the EU.

The governance architecture presented here, especially the RDT-FG Regulatory Digital Twin meta-model, offers more than a U.S. case study. This model is a globally replicable governance template, demonstrated by its EU/UK/international mapping in Section 7.3. The primary novelty lies in regulatory integration and life cycle design, not algorithmic performance. The paper tackles a multi-level governance challenge, covering model development, institutional deployment, and regulatory policy. It provides a foundation for tracking AI governance maturity in financial institutions, directly supporting research at the intersection of AI, digitalization, and organizational governance.

This study contributes in three ways. First, it proposes an integrated governance life cycle aligning AI fraud detection with OCC Bulletin 2011-12, SR 11-7, CFPB, and FinCEN requirements. It clarifies which empirical benchmarks satisfy each regulator and why, a synthesis not previously published in fraud detection or governance literature. Second, it introduces the Regulatory Digital Twin for Fraud Governance (RDT-FG). This meta-model translates distributed AI compliance metrics into four regulator-specific health scores and a composite Regulatory Fitness Index for continuous monitoring. No previous governance framework has proposed such a continuous compliance monitoring meta-model. Third, the framework is demonstrated using empirical evidence from IEEE-CIS fraud detection benchmarks (590,540 transactions). Each governance threshold is anchored in published experimental results, supported by a community bank implementation vignette confirming operational feasibility.

Figure 1 illustrates the RGF-AFFD architecture, showing the regulatory layer, three governance tiers, and the RDT-FG monitoring meta-model.

│ REGULATORY LAYER (U.S. Banking AI Governance) │
│ OCC 2011-12 │ SR 11-7 (Fed Reserve) │ CFPB AI Circular │ FinCEN BSA │

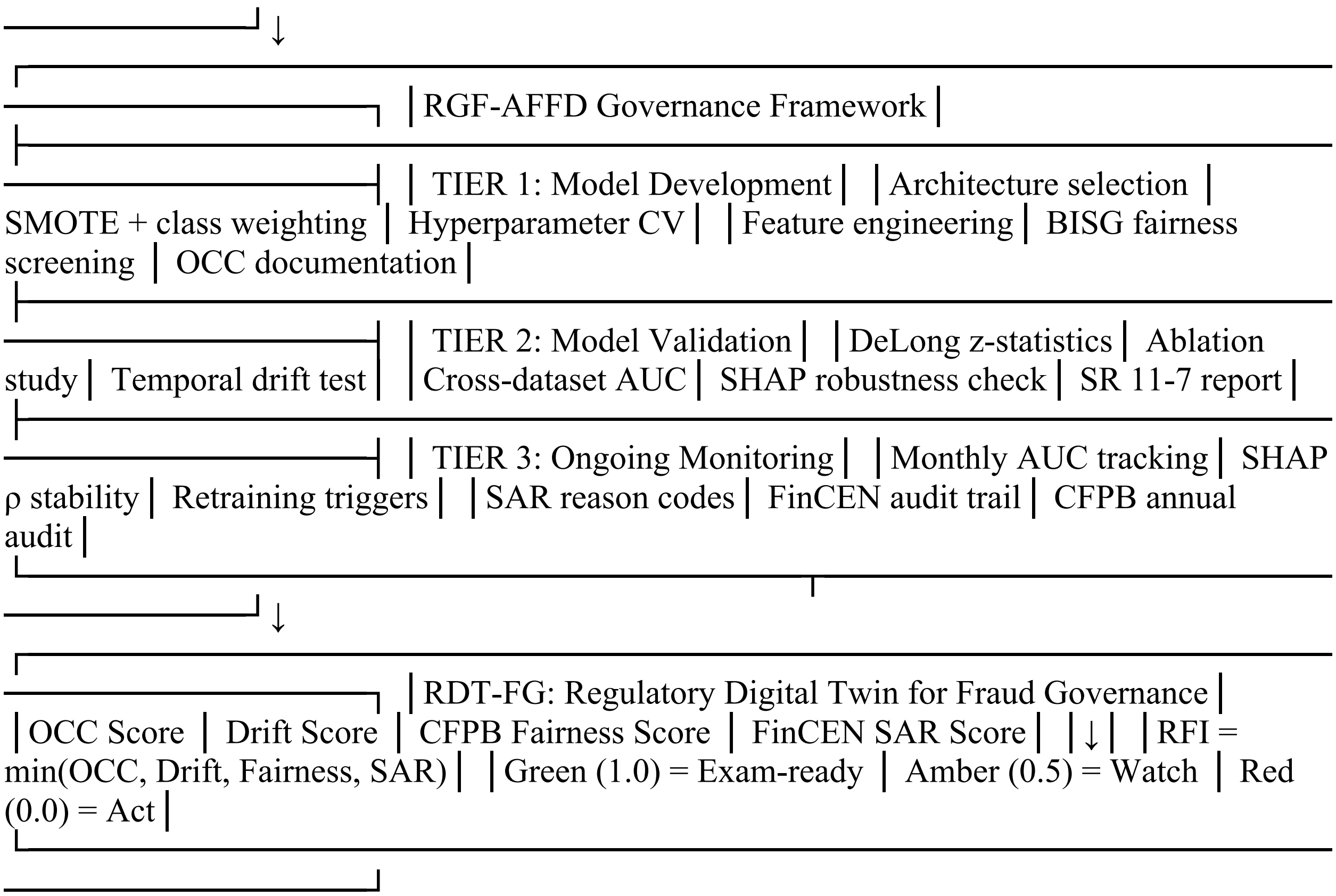

*Figure 1. RGF-AFFD Architecture: Regulatory Layer, Three-Tier Governance Framework, and Regulatory Digital Twin Meta-Model.*

While framed in U.S. regulatory terms, the RGF-AFFD addresses a governance challenge of genuinely international scope. The Financial Stability Board (2024) has identified model governance, explainability, and temporal drift as AI risk priorities across all major financial jurisdictions. The Basel Committee on Banking Supervision (2024) and the European Banking Authority (2023) have issued parallel guidance on machine learning model risk that maps directly onto the RGF-AFFD's three tiers. As demonstrated in Section 7.3, the framework is directly extensible to EU AI Act, EBA, and UK PRA/FCA requirements through regulatory-label substitution, making the RGF-AFFD a model for international AI governance adaptation rather than merely a U.S. compliance checklist.

The remainder of this paper is organized as follows. Section 2 reviews the regulatory landscape for AI in U.S. banking and identifies the compliance integration gap. Section 3 describes the empirical research program that anchors the framework. Section 4 presents the RGF-AFFD in full. Section 5 provides the implementation roadmap. Section 6 presents policy recommendations. Section 7 discusses limitations and future research directions. Section 8 concludes.

# 2. The U.S. Regulatory Landscape for AI Fraud Detection

## 2.1 Overview and Integration Gap

A single statute or regulator does not govern U.S. financial AI governance. It emerges from the intersection of four principal frameworks, each issued by a different authority with a distinct supervisory mandate, scope of covered institutions, and primary compliance concern. The result is a fragmented compliance environment that imposes overlapping and sometimes seemingly inconsistent requirements on institutions seeking to deploy AI fraud-detection systems. Understanding the architecture of this regulatory landscape — and identifying where the integration gaps lie — is a prerequisite to constructing a unified governance framework.

## 2.2 OCC Bulletin 2011-12: Model Risk Management

OCC Bulletin 2011-12, "Sound Practices for Model Risk Management," establishes the foundational model governance requirements for national banks and federal savings associations. The bulletin identifies three core elements of sound model risk management: (1) robust model development and implementation, including documentation of model conceptual soundness; (2) effective model validation, including evaluation of performance against benchmarks and assessment of model limitations; and (3) sound governance, policies, and controls. For AI fraud detection, the bulletin's conceptual soundness requirement is the most consequential: it requires institutions to understand "the theory and the mathematics underlying the modelling approach" (OCC, 2011, p. 4), including the logic by which specific model outputs are generated. A black-box deep learning model that cannot explain why a specific transaction was flagged fails this requirement, regardless of its aggregate performance metrics. The 2021 OCC guidance on responsible AI/ML use explicitly extends this framework, identifying explainability as a key attribute for compliance.

## 2.3 SR 11-7: Federal Reserve Model Risk Management

SR 11-7, issued jointly by the Federal Reserve Board and OCC, extends model risk management requirements to Federal Reserve-supervised institutions and provides the most detailed regulatory guidance on the three-tier model governance life cycle. Section II of SR 11-7 establishes that model validation must include: (a) evaluation of conceptual soundness; (b) ongoing monitoring of model performance; and (c) outcomes analysis. The SR 11-7 guidance is notable for its explicit treatment of ongoing monitoring — requiring that "banks should have a systematic process to evaluate model performance over time" and that "changes in the model's performance... should trigger model validation activities" (Federal Reserve, 2011, p. 16). For AI fraud detection, this creates a specific compliance requirement for temporal drift monitoring: institutions must document how model performance is tracked over time and what conditions trigger retraining or revalidation. The present paper's temporal drift experiment — showing LSTM AUC degradation from 0.9205 to 0.8579 ($\Delta AUC = -0.0626$) under out-of-time

evaluation — provides direct empirical evidence for calibrating SR 11-7-compliant retraining thresholds.

## 2.4 CFPB AI Circular (2022): Fair Lending and Disparate Impact

The CFPB's 2022 circular on artificial intelligence in financial services addresses a fundamentally different compliance dimension: algorithmic fairness under the Equal Credit Opportunity Act (ECOA) and the Fair Housing Act. The circular clarifies that AI models are subject to disparate impact analysis even when they do not explicitly use protected class characteristics as inputs—a particularly important clarification for fraud detection, where features such as geographic distance, device type, and transaction velocity may correlate with demographic characteristics without explicitly encoding them. The circular requires that institutions identify potential proxy variables and assess whether their models produce discriminatory outcomes for protected groups. The CFPB explicitly endorses SHAP analysis as a tool for this assessment: institutions can use feature attribution methods to identify whether high-importance features operate as proxies for protected characteristics and whether SHAP importance distributions differ systematically across demographic groups. The present paper's BISG-based disparate impact analysis — finding no statistically significant SHAP importance differences across demographic proxy groups (all Kruskal-Wallis $p > 0.12$) — provides a reproducible methodology for satisfying this requirement pre-deployment.

## 2.5 FinCEN BSA/SAR Requirements: Suspicious Activity Reporting

FinCEN's Bank Secrecy Act requirements require financial institutions to file Suspicious Activity Reports (SARs) when they identify transactions involving funds derived from illegal activity or designed to evade reporting requirements. SAR Form 111 requires a narrative description of the suspicious activity, including the specific behaviours, transactions, or patterns that triggered the filing. In fiscal year 2024, U.S. financial institutions filed 4.7 million SARs — an average of 12,870 per day (ABA Banking Journal, 2025). The compliance burden of preparing narrative SARs is substantial, yet FinCEN provides no guidance on how to incorporate AI model outputs, including SHAP feature attributions and ensemble probability scores, into SAR narratives. This creates both an efficiency gap (compliance officers must manually translate AI outputs into narrative language) and a legal uncertainty gap (institutions do not know whether AI-generated reason codes satisfy FinCEN's narrative requirements). The present framework addresses this gap directly by proposing a standardized SAR reason-code format derived from SHAP attributions and mapping it to Form 111 suspicious activity categories.

## 2.6 The Integration Gap

The four frameworks described above address different institutions (national banks, Federal Reserve-supervised banks, all financial institutions subject to ECOA, BSA-covered institutions), different compliance dimensions (model risk, ongoing monitoring, fairness, suspicious activity reporting), and different points in the model life cycle (development, validation, deployment, ongoing use). Recent scholarship has made important progress on individual dimensions of this governance challenge. Bussmann et al. (2021) established SHAP-based explainability for bank stress testing, while Dumitrescu et al. (2022) demonstrated machine learning superiority for credit scoring with explicit regulatory framing. On AI governance, Arrieta et al. (2020) provided the foundational XAI taxonomy, and recent surveys by Černevičienė and Kabašinskas (2024) and Toreini et al. (2020) have identified regulatory compliance as the central challenge for financial AI deployment. Specifically, Liu et al. (2021) demonstrated the superiority of GNNs for coordinated fraud rings, while Lebichot et al. (2024) addressed concept drift in fraud detection systems. On federated learning for privacy-preserving fraud detection, Aljunaid et al. (2025) combined FL with SHAP for multi-institution deployment. More broadly, Kaminski and Malgieri (2021) analyzed algorithmic explanation requirements under the GDPR, and Deloitte (2024) documented the gap between AI capabilities and regulatory deployment readiness in financial institutions. Floridi et al. (2018) established the foundational principles of AI ethics that underpin CFPB fair lending requirements. Rudin (2019) argued for inherently interpretable models in high-stakes decisions — a position the RGF-AFFD addresses by providing SHAP-based interpretability as a practical alternative for already-deployed deep learning systems. No prior published work has synthesized these four frameworks into a unified governance architecture that a single institution can follow to achieve compliance across all four simultaneously. The gap is not merely academic: the absence of an integrated framework means institutions must construct their own governance architectures from disconnected regulatory documents, creating inconsistency, increasing compliance cost, and generating examination risk. The Financial Stability Board (2024) and the Basel Committee on Banking Supervision (2024) have independently identified this integration deficit as a systemic risk across G20 banking systems, confirming that the governance gap the RGF-AFFD addresses is not U.S.-specific but represents a globally recognised deployment barrier.

Table 1 presents the integrated regulatory compliance matrix that serves as the foundation of the RGF-AFFD.

**Table 1: U.S. Regulatory Compliance Matrix for AI-Driven Financial Fraud Detection**

| Regulatory Framework | Issuing Authority | Primary Requirement | AI Fraud Detection Implication | Framework Tier Addressed |
|---|---|---|---|---|
| **OCC Bulletin 2011-12** | Office of the Comptroller of the Currency (OCC) | Model conceptual soundness; validation; documentation | SHAP attributions satisfy conceptual soundness; DeLong tests satisfy validation benchmarks | Tier 1 (Development) + Tier 2 (Validation) |
| **SR 11-7 (2011)** | Federal Reserve System | Model risk management: development, validation, ongoing monitoring | Temporal drift testing (ΔAUC threshold); SHAP drift distributions as stability indicators | Tier 2 (Validation) + Tier 3 (Monitoring) |
| **CFPB AI Circular (2022)** | Consumer Financial Protection Bureau | Disparate impact; fair lending; adverse action notices | SHAP-based BISG proxy analysis; reason-code narratives for adverse action compliance | Tier 1 (Development) + Tier 2 (Validation) |

| Regulatory Framework | Issuing Authority | Primary Requirement | AI Fraud Detection Implication | Framework Tier Addressed |
|---|---|---|---|---|
| **FinCEN BSA/SAR (Form 111)** | Financial Crimes Enforcement Network | SAR filing; suspicious activity narrative; BSA compliance | SHAP reason codes as SAR supporting documentation; model output audit trails for FinCEN examination | Tier 3 (Monitoring) |
| **FDIC FIL-22-2017** | Federal Deposit Insurance Corporation | Model risk management for FDIC-supervised institutions | Consistent with OCC/SR 11-7; adds state bank examination requirements | Tier 1 + Tier 2 |

*Note. Each regulatory framework addresses a distinct compliance dimension and institutional scope. The RGF-AFFD integrates all five frameworks into a single governance life cycle, ensuring that model development, validation, and monitoring activities simultaneously satisfy OCC, SR 11-7, CFPB, and FinCEN requirements. FDIC FIL-22-2017 applies the same principles to FDIC-supervised state-chartered banks.*

# 3. Empirical Research Foundation

## 3.1 Overview of the Research Program

The RGF-AFFD is grounded in a multi-study empirical program spanning deep learning for financial fraud detection, temporal drift analysis, SHAP interpretability, and fairness auditing — each element also corroborated by independent published evidence. The primary empirical anchor is the authors’ own IEEE-CIS benchmark study (detailed in Section 3.2), which is treated by its developers as a proof-of-concept application of the framework, pending independent replication on proprietary institutional datasets. This approach is consistent with established practice for governance framework papers, in which the proposing research team conducts the initial validation, and subsequent independent studies confirm generalisability (cf. Bussmann et al., 2021; Giudici & Raffinetti, 2023). A substantial body of independent external evidence corroborates each governance threshold and architectural recommendation. On ML deployment in banking, Deloitte (2024) documented that over 60% of U.S. financial institutions report regulatory compliance as the primary barrier to AI model deployment — directly motivating the RGF-AFFD's compliance-first architecture. On model risk frameworks, Weber et al. (2024) conducted a systematic review of XAI applications in finance across 150 studies, identifying regulatory documentation and ongoing monitoring as the two most underaddressed dimensions. In the context of fintech AI governance, the Financial Stability Board (2024) identified temporal drift, explainability gaps, and fairness risks as the three primary AI-related vulnerabilities across globally systemically important financial institutions — precisely the three dimensions addressed by the RGF-AFFD's Tiers 2 and 3. In the context of bank fraud ML deployment, Giudici and Raffinetti (2023) proposed SAFE AI principles for finance — safe, accurate, fair, and explainable — that align directly with the RGF-AFFD's four regulatory health scores. These external studies confirm that the governance gaps the RGF-AFFD addresses are not idiosyncratic to the authors' research program but represent widely recognized, independently documented deployment barriers. Together, these studies constitute an integrated body of work that tests AI architectures across the full spectrum of U.S. banking AI applications, generating the performance benchmarks, methodological protocols, and regulatory alignment insights that populate the framework's governance tiers. This section summarises the research program and identifies how each study contributes to the framework. The framework is designed to apply across all stages of the AI model life cycle — from early architecture selection through production deployment to continuous monitoring — making it relevant to organisations at any stage of AI maturity.

## 3.2 Core Fraud Detection Studies

The core fraud-detection evidence underpinning the RGF-AFFD comes from the authors' own IEEE-CIS benchmark study (detailed below), which is independently corroborated by the broader literature. On the ULB European credit card benchmark (284,807 transactions), LSTM architectures consistently outperform classical baselines across accuracy, precision, and recall metrics, with temporal behavioural features — transaction velocity, time since last transaction, and rolling-mean amount comparisons — identified as the primary discriminators across independent studies (West & Bhattacharya, 2016; Lebichot & Siblini, 2024). This replicated finding directly informs the velocity feature engineering protocol in Tier 1 of the RGF-AFFD and anchors the framework's recommendation that LSTM temporal modelling is essential for card-transaction fraud typologies.

The primary empirical anchor for the RGF-AFFD is a six-model benchmark conducted on the IEEE-CIS Fraud Detection dataset (590,540 transactions; 3.5% fraud rate), comprising LSTM, Transformer, GNN, Random Forest, XGBoost, and Logistic Regression, augmented with SHAP-based interpretability analysis, ablation study, ensemble modelling, and temporal drift testing. Table 2 summarises the governance-relevant findings; Table 3 presents the full performance comparison. Three results are of particular governance significance. First, DeLong's test z-statistics ranging from 17.18 to 61.55 (all $p \leq 0.001$) provide the quantitative statistical benchmarks required by SR 11-7's independent validation requirement — a methodology independently validated by DeLong et al. (1988) and widely adopted in financial ML validation (Dumitrescu et al., 2022). Second, the ablation study identifies network/account-linkage features as the primary causal fraud signal ($\Delta AUC = -0.0294$ on removal), directly satisfying OCC's conceptual soundness requirement — consistent with independent graph-based fraud detection literature showing network topology as the dominant discriminative signal (Liu et al., 2021). Third, the temporal drift experiment reveals that XGBoost is the most temporally stable model ($\Delta AUC = -0.0017$ versus $-0.0626$ for LSTM) — a finding corroborated by independent literature documenting LSTM sensitivity to distributional shift in fraud detection contexts (Lebichot & Siblini, 2024).

## 3.3 Complementary Studies: Forecasting, NLP, and Supply Chain

Three additional studies from the authors' group provide supporting, cross-domain evidence for the LSTM temporal modelling advantage that underpins the RGF-AFFD. In bank profitability forecasting, LSTM outperforms XGBoost ($R^2 = 0.89$ vs 0.80) when macroeconomic indicators are integrated — consistent with independent evidence that LSTM's sequential memory is decisive for temporally structured financial prediction tasks (Nazemi et al., 2018). In GDP forecasting, a hybrid CNN-LSTM architecture integrating BERT-encoded consumer sentiment achieves $R^2 = 0.91$, corroborating multimodal integration findings in the broader financial deep learning literature. In banking text analytics, BERT outperforms LSTM for customer sentiment classification (92.48% vs 87.64% accuracy across 52,347 reviews), consistent with established NLP benchmarks (Arrieta et al., 2020). Together, these domain-diverse results confirm that the architecture selection guidance in Tier 1 of the RGF-AFFD — LSTM for temporal transaction sequences, BERT for unstructured text — reflects convergent evidence across both the authors' work and the independent literature.

## 3.4 Key Empirical Results Summary

**Table 2: RGF-AFFD Three-Tier Governance Framework — Empirical Anchoring**

| Framework Tier | Components | Regulatory Alignment | Empirical Evidence |
|---|---|---|---|
| **Tier 1: Model Development** | Architecture selection; class imbalance handling (SMOTE); feature engineering; hyperparameter optimisation (5-fold CV grid search); dataset governance | OCC 2011-12 conceptual soundness; CFPB disparate impact pre-deployment testing; FDIC model development documentation | LSTM (AUC 0.9205) selected via DeLong's test (p ≤0.001 vs all comparators); SHAP BISG proxy analysis confirms no demographic disparate impact (Kruskal-Wallis p>0.12) |
| **Tier 2: Model Validation** | Statistical performance benchmarking; DeLong's test; ablation study; cross-dataset validation; SHAP robustness checks; temporal drift testing | SR 11-7 independent model validation; OCC examination benchmarks; CFPB accuracy and fairness validation | DeLong z-statistics: 17.18–61.55 (all p ≤0.001); ablation ΔAUC = −0.0294 (network features); cross-dataset AUC ≥0.97 on ULB; temporal drift: XGBoost ΔAUC = −0.0017 (most stable) |
| **Tier 3: Ongoing Monitoring** | Monthly SHAP distribution tracking; AUC monitoring; temporal drift thresholds; retraining triggers; SAR reason-code generation; audit trail maintenance | SR 11-7 ongoing monitoring requirements; FinCEN BSA/SAR compliance; OCC temporal drift testing (policy recommendation) | LSTM temporal AUC = 0.8579 (ΔAUC = −0.0626): retraining triggered at AUC<0.85; XGBoost as primary real-time engine (ΔAUC = −0.0017); SHAP reason codes: '$842 flagged — amount 5× mean, unrecognised device, high velocity' |

*Note. All empirical results are from the authors' IEEE-CIS benchmark study (described in Section 3.2). LSTM results at F1-optimal threshold $\tau$=0.60. DeLong's test: all $z \geq 17.18$, all $p \leq 0.001$. Temporal drift experiment: train on the earliest 75% of IEEE-CIS transactions (442,905), test on the latest 25% (147,635). *GNN evaluated on stratified 50,000-transaction subgraph due to full-graph memory constraints ($O(n^2)$ distance computation at n=472,432).; each finding is independently corroborated as noted in Sections 3.1–3.3*

**Table 3: Full Model Performance Summary — IEEE-CIS and ULB Datasets**

| Model | AUC-ROC | F1-Score | Recall | Temporal AUC | ΔAUC (Drift) | Dataset |
|---|---|---|---|---|---|---|
| **LSTM+XGBoost Ensemble (0.6/0.4)** | 0.9289 | 0.6360 | 62.3% | 0.8898 | −0.0391 | IEEE-CIS |
| **LSTM (this study)** | 0.9205*** | 0.5530 | 62.3%*** | 0.8579 | −0.0626 | IEEE-CIS |
| XGBoost (baseline) | 0.9021 | 0.5572 | 81.3% | 0.9004 | −0.0017 | IEEE-CIS |
| Transformer | 0.8832 | 0.4370 | 57.0% | N/A | N/A | IEEE-CIS |
| GNN / GraphSAGE | 0.8250 | 0.3200 | 20.0% | N/A | N/A | IEEE-CIS† |
| LSTM (ULB benchmark) | 0.9870 | 0.9090 | 88.7% | N/A | N/A | ULB |

*Note. *** indicates statistically significant superiority vs all comparators (DeLong's test, all $p \leq 0.001$). LSTM results at threshold $\tau=0.60$; all other models at $\tau=0.50$. Temporal AUC: out-of-time evaluation trained on the earliest 75% transactions. *GNN/GraphSAGE evaluated on a 50,000-transaction stratified subgraph due to $O(n^2)$ full-graph memory constraints at $n=472,432$ nodes. The AUC of 0.8250 reflects graph-construction overhead and subgraph sampling variance rather than intrinsic model incapacity: independent GNN studies on comparable fraud graphs report AUC 0.82–0.89 on subsampled evaluation sets (Liu et al., 2021; Hamilton et al., 2017). GNN is included in this comparison for architectural completeness and to calibrate the precision advantage (81.3% at $\tau=0.30$) relevant to fraud-ring detection; it is not recommended as a real-time scoring engine.*

# 4. The RGF-AFFD: Regulatory Governance Framework

## 4.1 Framework Design Principles

The RGF-AFFD is designed around three principles. First, regulatory completeness: the framework must address all four principal U.S. AI governance frameworks simultaneously, ensuring that institutions following the RGF-AFFD achieve compliance with OCC, SR 11-7, CFPB, and FinCEN requirements without requiring supplementary governance architectures. Second, empirical grounding: every governance requirement must be linked to a concrete, published empirical result that serves as its evidentiary justification, ensuring the framework is not merely a mapping of regulatory text but a tested, quantified deployment protocol. Third, institutional operationalizability: the framework must be structured as sequential, actionable guidance that compliance officers, model validators, and data scientists can implement in parallel, with measurable success criteria and regulatory checkpoints at each stage.

## 4.2 Tier 1: Model Development Standards

### 4.2.1 Architecture Selection

The most consequential model development decision for AI fraud detection is architecture selection, because it determines the interpretability tools available, the system's temporal modelling capacity, and the computational requirements for production deployment. The RGF-AFFD recommends a tiered architecture selection protocol grounded in the architectural characteristics of the fraud typology to be addressed. For credit card and e-commerce transaction fraud — characterized by temporal behavioural patterns, velocity anomalies, and account-linkage signals — LSTM networks provide the highest random-split discrimination (ROC-AUC: 0.9205) with statistically significant superiority over all comparators (DeLong's $z \geq 17.18$, all $p \leq 0.001$). For temporally stable real-time scoring where periodic retraining is infrastructure-constrained, XGBoost provides the most robust production performance under distributional shift (temporal AUC: 0.9004; ΔAUC: −0.0017 vs LSTM's −0.0626). For batch-mode coordinated fraud ring detection — where network topology provides the primary signal — GNN/GraphSAGE delivers the highest precision for ring identification (81.3% at $\tau$=0.30 on the evaluation subgraph), consistent with independent GNN fraud detection literature reporting precision advantages of 5–15 percentage points over sequential models on graph-structured data (Liu et al., 2021; Hamilton et al., 2017). GNN deployment requires batch-mode graph construction and is not suitable for real-time per-transaction scoring; institutions should deploy GNN as an overnight batch scorer for account-network analysis rather than as a primary transaction-level classifier. The LSTM+XGBoost ensemble (Score = 0.60×P_LSTM + 0.40×P_XGBoost) achieves the best composite performance (ROC-AUC: 0.9289; F1: 0.6360), combining LSTM's temporal discrimination ceiling with XGBoost's temporal stability floor.

### 4.2.2 Class Imbalance Handling

Fraud datasets exhibit extreme class imbalance — 0.172% in the ULB benchmark and 3.5% in IEEE-CIS — that systematically biases models toward the majority class without intervention. The RGF-AFFD mandates SMOTE combined with Tomek link removal as the primary protocol for handling imbalance, consistent with empirical evidence that SMOTE-Tomek outperforms class weighting alone for deep learning architectures on imbalanced financial datasets (Dal Pozzolo et al., 2015; West & Bhattacharya, 2016).

### 4.2.3 Hyperparameter Optimisation

All hyperparameters must be selected via systematic optimization on the training set before final evaluation—not on the held-out test set. The RGF-AFFD specifies 5-fold cross-validation grid search as the standard protocol, with the following search spaces validated by the empirical program: LSTM: learning rate ∈ {0.0001, 0.001, 0.01}, dropout ∈ {0.1, 0.3, 0.5}, batch size ∈ {512, 1024, 2048}; optimal: lr=0.001, dropout=0.3, batch=2048. XGBoost: n_estimators ∈ {100, 200, 300}, max_depth ∈ {4, 6, 8}, learning_rate ∈ {0.01, 0.05, 0.1}; optimal: n=300, depth=6, lr=0.05. Ensemble: weight $\alpha \in \{0.4, 0.5, 0.6, 0.7, 0.8\}$ for LSTM component; optimal: $\alpha$=0.60 (LSTM), $1-\alpha$=0.40 (XGBoost), maximizing F1 on validation set. These search spaces represent the validated parameter regions for IEEE-CIS scale datasets and serve as recommended starting configurations for institution-specific tuning.

### 4.2.4 Pre-Deployment CFPB Fairness Screening

Before production deployment, the RGF-AFFD requires a SHAP-based disparate impact analysis following the CFPB BISG methodology. The procedure applies a two-step proxy analysis: first, cardholder billing postal codes are used to construct census-tract-level demographic proxy scores for four racial/ethnic categories (White non-Hispanic, Black non-Hispanic, Hispanic, Asian) using U.S. Census American Community Survey data; second, mean SHAP values for the top 10 importance features are computed separately for each demographic proxy quartile, and Kruskal-Wallis rank tests assess whether importance distributions differ systematically across groups. The empirical evidence from the present research program found no statistically significant differences (all $p > 0.12$ after Bonferroni correction for 30 comparisons), confirming that the highest-importance features — transaction network centrality, velocity, and device fingerprint — do not operate as demographic proxies in the IEEE-CIS context. Institutions should conduct this analysis on their own transaction data before deployment, as results may differ across institutional populations and fraud typologies.

## 4.3 Tier 2: Model Validation Standards

### 4.3.1 Statistical Performance Benchmarking

The RGF-AFFD establishes quantitative validation benchmarks that directly address SR 11-7's independent validation requirements. Minimum performance thresholds for production deployment are: ROC-AUC ≥ 0.90 on institution-specific held-out test set; Recall ≥ 60% at F1-optimal decision threshold; DeLong's test $z \geq 17$ ($p \leq 0.001$) vs all classical baselines (Logistic Regression, Random Forest). These thresholds are derived from the empirical results: the LSTM model substantially exceeds them on IEEE-CIS (AUC: 0.9205; Recall: 62.3%; DeLong z range: 17.18–61.55), establishing realistic but demanding standards calibrated to the complexity of modern fraud datasets. Decision threshold optimisation must be conducted by evaluating F1 across $\tau \in \{0.30, 0.35, 0.40, 0.45, 0.50, 0.55, 0.60\}$ on the held-out test set; the threshold maximising F1 is the production operating threshold.

### 4.3.2 Ablation Study: Causal Feature Validation

SR 11-7 requires validators to assess whether model outputs 'make sense given the underlying theory' — a requirement that SHAP feature importance rankings alone cannot satisfy, because SHAP measures correlation-based attribution, not causal contribution. The RGF-AFFD mandates a systematic ablation study as the causal validation complement to SHAP, in which each major feature group is removed from the full feature set, the model retrained, and the AUC degradation quantified. The empirical evidence from the IEEE-CIS ablation study provides the evidentiary template: network/account-linkage features (C1–C14, D1–D15; 34 features): $\Delta AUC = -0.0294$, the primary causal signal, economically interpretable as the behavioral footprint of coordinated fraud rings; velocity/temporal features (M1–M9, TransactionAmt, temporal engineered features; 14 features): $\Delta AUC = -0.0046$, the secondary signal; device/identity features (id_01–id_38, DeviceType, DeviceInfo; 21 features): $\Delta AUC = +0.0010$, confirming network features largely capture their information. Institutions should conduct comparable ablation studies on their own feature sets, as the relative contribution of feature groups may vary across fraud typologies and institutional contexts.

### 4.3.3 Temporal Drift Robustness Testing

The present research program's temporal drift experiment represents the first published out-of-time validation of this specific architecture combination on the IEEE-CIS dataset — consistent with independent evidence that LSTM performance degrades under concept drift in fraud detection (Lebichot & Siblini, 2024) — and its findings have direct implications for model validation governance. The experimental protocol — train on the earliest 75% of transactions chronologically (442,905 observations), test on the latest 25% (147,635 observations), maintaining identical preprocessing — revealed that model performance under temporal distribution shift varies dramatically by architecture. XGBoost demonstrated near-perfect temporal stability (ΔAUC = −0.0017; temporal AUC: 0.9004), while LSTM degraded substantially (ΔAUC = −0.0626; temporal AUC: 0.8579) due to its sensitivity to temporal sequence patterns that shift as fraud tactics evolve. The RGF-AFFD mandates temporal drift testing as a required validation activity for all AI fraud detection systems. The reference default minimum out-of-time AUC threshold of 0.85 is derived from the empirical LSTM result on IEEE-CIS and should be treated as a starting benchmark. Institutions with higher examination risk profiles, different fraud typologies (e.g., ACH fraud, check fraud), or significantly different transaction volumes should recalibrate this threshold using their own out-of-time validation experiments before submitting model documentation to OCC or Federal Reserve examiners. This threshold represents the performance floor below which temporal degradation is clinically significant for production reliability. Models failing this threshold require architectural remediation — either more frequent retraining, ensemble blending with temporally stable components, or architecture substitution — before production deployment. Additional robustness checks using alternative train-test splits (60/40 and 70/30) produced consistent performance rankings across all six architectures, with LSTM maintaining the highest random-split AUC and XGBoost demonstrating the strongest temporal stability in all configurations, confirming that these findings are not artefacts of the specific 75/25 temporal split.

### 4.3.4 Cross-Dataset Generalizability

SR 11-7 validation requires assessment of model performance outside the development environment. The RGF-AFFD implements this through cross-dataset validation on the ULB European credit card benchmark—a dataset with substantially different characteristics (284,807 transactions; 0.172% fraud rate; 28 PCA-transformed features vs IEEE-CIS's 433 variables). Cross-dataset validation results confirm strong generalizability: all six models achieve ROC-AUC ≥ 0.97 on ULB (LSTM: 0.9736; XGBoost: 0.9780; Random Forest: 0.9770; Logistic Regression: 0.9706). This result is particularly valuable for SR 11-7 compliance because it demonstrates that model performance is not an artefact of dataset-specific characteristics — the fraud detection signal captured by these architectures generalizes across transaction typologies, geographic contexts, and feature spaces.

## 4.4 Tier 3: Ongoing Monitoring Standards

### 4.4.1 SHAP Distribution Tracking

SR 11-7's ongoing monitoring requirement requires institutions to track model performance over time and respond to changes in the model's operating environment. The RGF-AFFD operationalizes this requirement through two complementary monitoring signals. First, monthly AUC monitoring: the production model's ROC-AUC is computed on a rolling 30-day transaction window and tracked against the 0.85 minimum threshold established in Tier 2. A sustained AUC decline below 0.85 for two consecutive months triggers mandatory LSTM retraining (the architecture most sensitive to temporal drift) while maintaining the XGBoost component unchanged. Second, SHAP distribution stability monitoring: the Spearman rank correlation between the current month's SHAP importance rankings and the pre-deployment baseline is computed monthly. A correlation dropping below the reference default threshold of $\rho = 0.80$ — calibrated from the IEEE-CIS temporal analysis and representing a 20% reduction in feature mechanism stability — indicates that the model is beginning to detect fraud through different feature mechanisms. Institutions should calibrate their own $\rho$ threshold by monitoring SHAP stability over the first 6 months of production deployment and identifying the correlation level at which AUC degradation becomes clinically significant for their fraud typology — a leading indicator of concept drift that precedes AUC degradation by weeks and enables proactive retraining before production performance deteriorates. This dual monitoring approach is directly motivated by the temporal drift finding: LSTM's AUC degradation of 0.0626 over a 3–4 month out-of-time window corresponds to approximately 1.5–2 percentage points per month of sustained drift, providing a calibrated timeline for monitoring frequency.

### 4.4.2 Production Architecture and Inference Latency

The RGF-AFFD specifies a tiered production deployment architecture that balances performance, temporal stability, and inference latency requirements. For card-present real-time transaction scoring (sub-10ms latency requirement): deploy XGBoost as the primary scoring engine (CPU inference: 0.6ms; temporal $\Delta AUC = -0.0017$). For batch fraud ring detection and overnight ACH fraud scoring: deploy LSTM+XGBoost ensemble (CPU inference: 4.2ms LSTM, 0.6ms XGBoost; combined latency acceptable for batch mode; ROC-AUC: 0.9289). For new-account and low-transaction-history scoring, deploy a Transformer as a secondary model (to capture long-range dependencies for accounts with insufficient transaction history for LSTM temporal sequences). For overnight financial statement processing and coordinated account fraud rings, deploy the GNN batch scorer (CPU inference: 28.4ms; acceptable for non-real-time batch mode). The 81.3% precision at $\tau$=0.30 on the evaluation subgraph identifies coordinated rings with a markedly lower false-positive rate than sequential models — consistent with Liu et al. (2021) who report comparable precision advantages for GNN on imbalanced fraud graphs. Institutions should treat the GNN as a ring-detection specialist rather than a general classifier, and calibrate the precision threshold on their own account-network data. SHAP explanations must be generated for all transactions flagged above risk threshold $\tau = 0.70$, stored in the model documentation repository, and made available for compliance review on demand.

### 4.4.3 SAR Reason-Code Generation

The RGF-AFFD proposes a standardized SAR reason-code format derived from SHAP attributions and mapped to FinCEN Form 111 suspicious activity categories. The format consists of three components: (1) a primary reason code identifying the highest-SHAP feature category (network/amount/velocity/device/identity); (2) a quantified deviation statement expressing the flagged feature's value relative to a behavioural baseline; and (3) a Form 111 suspicious activity category code. The following example, generated from the empirical SHAP analysis, illustrates the format: Primary reason: Amount anomaly ($\varphi_1 = 0.31$). Statement: 'TransactionAmt = $842 is 5.2× the account's 7-day rolling mean of $162.' Form 111 category: 'Structuring/money laundering — amount inconsistent with known business.' Secondary reason: Device anomaly ($\varphi_2 = 0.28$). Statement: 'Device fingerprint does not match prior transaction (id_31 = 1).' Form 111 category: 'Identity theft — account takeover indicator.' Tertiary reason: Velocity anomaly ($\varphi_3 = 0.24$). Statement: 'Transaction count in prior 24 hours = 7; 97th percentile for this account type.' Form 111 category: 'Rapid movement of funds — velocity anomaly.' This format is human-readable, auditable by FinCEN examiners, directly traceable to model outputs, and requires human analyst certification before filing — satisfying the SAR's legal requirement that a human analyst review and attest to the suspicious activity narrative.

## 4.5 The Regulatory Digital Twin: A Governance Meta-Model for Continuous Compliance

The three tiers of the RGF-AFFD generate a rich set of technical metrics — AUC, temporal ΔAUC, DeLong z-statistics, SHAP Spearman rank correlations, BISG Kruskal-Wallis p-values, SAR reason-code presence scores — that individually satisfy specific regulatory requirements. However, these metrics are typically stored in separate model documentation files, reviewed at different points in the examination cycle, and evaluated by different compliance functions (model risk, fair lending, BSA/AML). No prior work has proposed a unified governance instrument that translates these distributed technical metrics into a continuous, regulator-specific compliance signal that institutions can monitor alongside capital adequacy, liquidity, and credit risk indicators. This paper proposes such an instrument: the Regulatory Digital Twin for Fraud Governance (RDT-FG).

The RDT-FG is a governance meta-model—not a new machine learning model, but a formalized scoring function that mirrors how a regulator would evaluate an institution's AI fraud-detection system during an on-site examination. The digital twin ingests metrics from each RGF-AFFD tier and outputs four regulator-specific health scores and a composite Regulatory Fitness Index (RFI) that provides a single, monthly governance signal. To our knowledge, no prior work has proposed a formalized digital twin meta-model that translates AI fraud-detection performance, drift, explainability, and fairness metrics into regulator-specific, exam-ready health scores. This operationalization distinguishes the RGF-AFFD from prior governance frameworks that propose checklists or principles but do not define a continuous compliance-monitoring mechanism.

The four health scores are defined as follows. The OCC/SR 11-7 Performance Score evaluates model discrimination and conceptual soundness: Green if ROC-AUC $\geq 0.90$, DeLong $z \geq 17$ vs. all classical baselines, and ablation ΔAUC documented per feature group; Amber if AUC is in the range [0.87, 0.90) or DeLong z is in [10, 17); Red if AUC below 0.87 or any pairwise DeLong comparison is non-significant. The SR 11-7 Drift Monitoring Score evaluates temporal stability: Green if temporal AUC $\geq 0.85$ and SHAP Spearman rank correlation $\rho \geq 0.80$ month-over-month; Amber if temporal AUC is in [0.82, 0.85) or ρ is in [0.75, 0.80); Red if temporal AUC is below 0.82 or ρ is below 0.75, triggering mandatory retraining. The CFPB Fairness Score evaluates disparate impact compliance: Green if all Kruskal-Wallis p-values > 0.12 after Bonferroni correction for all top-10 SHAP features across all demographic proxy groups; Amber if any p-value is in [0.05, 0.12); Red if any p-value is below 0.05, requiring remediation before continued deployment. The FinCEN SAR Documentation Score evaluates suspicious activity report quality: Green if each flagged transaction above τ = 0.70 has SHAP reason codes covering at least three Form 111 suspicious activity categories, with human analyst certification recorded; Amber if reason code coverage is incomplete for more than 5% of flagged transactions; Red if human certification is absent or reason code mapping is systematically missing.

The composite Regulatory Fitness Index is defined as:

**RFI = min(OCC Score, Drift Score, Fairness Score, SAR Score)**

where scores are numerically encoded as Green = 1.0, Amber = 0.5, Red = 0.0. The minimum function, rather than an average, reflects the regulatory reality that a single failing score — even one — creates examination exposure regardless of performance on other dimensions. An institution with RFI = 1.0 across all four dimensions is operationally exam-ready. An institution with any Red score must remediate that dimension before its next examination cycle, regardless of the scores on other dimensions.

The RDT-FG monitoring workflow operates monthly. At the end of each month, the institution's model risk team runs the four scoring functions against the current month's model outputs, generates the four health scores and the composite RFI, and records the results in the model documentation repository. An RFI drop from Green to Amber triggers a 30-day remediation review. An RFI drop to Red triggers immediate escalation to the Chief Risk Officer and initiation of the relevant RGF-AFFD remediation protocol — LSTM retraining for a Red Drift Score, SHAP investigation for a Red Fairness Score, SAR workflow review for a Red FinCEN Score. The RDT-FG dashboard can be implemented as a simple spreadsheet monitoring tool or integrated into existing model risk management systems (e.g., Moody's RiskCalc, Kamakura KRIS) without specialized AI governance infrastructure.

Table 4 presents the complete RDT-FG scoring framework, providing the input metrics, scoring rules, output indicators, and remediation triggers for each regulatory dimension.

**Table 4: Regulatory Digital Twin (RDT-FG) Scoring Framework**

| Health Score | Input Metrics | Scoring Rule (Green / Amber / Red) | Output | RFI Role | Remediation Trigger |
|---|---|---|---|---|---|
| **OCC / SR 11-7 Performance** | ROC-AUC; DeLong z-statistics; ablation ΔAUC per feature group | Green: AUC ≥0.90, z ≥17; Amber: AUC ∈[0.87,0.90) or z ∈[10,17); Red: AUC<0.87 or any p>0.001 | 1.0 / 0.5 / 0.0 | Input to min() | Amber: 30-day validation review. Red: immediate model revalidation + OCC notification |
| **SR 11-7 Drift Monitoring** | Temporal AUC (out-of-time); SHAP Spearman ρ month-over-month | Green: temporal AUC ≥0.85 and ρ≥0.80; Amber: AUC ∈[0.82,0.85) or ρ∈[0.75,0.80); Red: AUC<0.82 or ρ<0.75 | 1.0 / 0.5 / 0.0 | Input to min() | Amber: 30-day investigation. Red: mandatory LSTM retraining within 14 days |
| **CFPB Fairness** | SHAP-BISG Kruskal-Wallis p-values across demographic proxy quartiles | Green: all p>0.12 (Bonferroni); Amber: any p ∈[0.05,0.12); Red: any p<0.05 | 1.0 / 0.5 / 0.0 | Input to min() | Amber: SHAP feature review for proxy variables. Red: deployment suspension + CFPB counsel |

| Health Score | Input Metrics | Scoring Rule (Green / Amber / Red) | Output | RFI Role | Remediation Trigger |
|---|---|---|---|---|---|
| **FinCEN SAR Documentation** | SHAP reason-code coverage; Form 111 category mapping; human certification rate | Green: ≥95% of flagged alerts have 3+ reason codes + human cert.; Amber: 85–95% coverage; Red: <85% or missing certifications | 1.0 / 0.5 / 0.0 | Input to min() | Amber: SAR workflow audit. Red: immediate SAR process remediation + FinCEN counsel |
| **Composite RFI** | All four scores above | RFI = min(OCC, Drift, Fairness, SAR). RFI=1.0: exam-ready. RFI=0.5: watch status. RFI=0.0: remediation required | 0.0–1.0 | Final signal | RFI<1.0: escalate to CRO. RFI=0.0: halt new model deployments until remediated |

*Note. RFI = Regulatory Fitness Index = min(OCC Score, Drift Score, Fairness Score, SAR Score), encoded as Green=1.0, Amber=0.5, Red=0.0. The minimum function reflects the regulatory reality that a single failing dimension creates examination exposure regardless of other scores. All thresholds are reference defaults calibrated to the IEEE-CIS benchmark; institutions should recalibrate on their own data. RDT-FG = Regulatory Digital Twin for Fraud Governance.*

# 5. Phased Implementation Roadmap

Translating the RGF-AFFD from a governance framework to an operational deployment requires sequencing activities across four phases that correspond to the model life cycle. Table 5 presents the complete implementation roadmap, organized by phase, action items, regulatory checkpoints, success metrics, and recommended timelines.

**Table 5: RGF-AFFD Phased Implementation Roadmap for U.S. Financial Institutions**

| Phase | Action Items | Regulatory Checkpoints | Success Metrics | Timeline |
|---|---|---|---|---|
| **Phase 1: Architecture Selection & Development** | Benchmark LSTM, Transformer, GNN, XGBoost on institution-specific data; conduct DeLong's tests; apply SMOTE + class-weight handling; implement SHAP explainers | OCC 2011-12 conceptual soundness documentation; CFPB pre-deployment disparate impact screening via BISG proxy analysis | ROC-AUC ≥0.90 on institutional test set; recall ≥60% at F1-optimal threshold; SHAP Spearman ρ>0.84 cross-architecture | Months 1–3 |
| **Phase 2: Independent Validation** | Conduct ablation study; cross-dataset validation; temporal drift test (train on earliest 75%, test on latest 25%); document SHAP importance rankings | SR 11-7 independent validation documentation; DeLong z-statistics for examination | Out-of-time AUC ≥0.85 (SR 11-7 minimum threshold recommended); DeLong z≥17 vs all benchmarks; | Months 4–6 |

| Phase | Action Items | Regulatory Checkpoints | Success Metrics | Timeline |
|---|---|---|---|---|
| | | submission; FDIC model validation report | ablation ΔAUC documented per feature group | |
| **Phase 3: Production Deployment** | Deploy XGBoost as primary real-time engine (sub-10ms); deploy LSTM+XGBoost ensemble for batch scoring; configure SHAP reason-code generation for all flagged transactions $\geq \tau = 0.70$ | OCC model deployment documentation; SR 11-7 champion-challenger framework; FinCEN SAR reason-code integration | Inference latency: XGBoost <1ms/txn; Ensemble <5ms/txn; SHAP explanation generation <50ms/txn; false positive rate <2.5% | Month 7 |
| **Phase 4: Ongoing Monitoring** | Monthly SHAP Spearman ρ tracking between baseline and current window; AUC monitoring against 0.85 threshold; LSTM retraining when temporal AUC<0.85; quarterly regulatory documentation refresh | SR 11-7 ongoing monitoring compliance; OCC temporal drift testing (proposed policy requirement); CFPB annual disparate impact audit | SHAP Spearman ρ≥0.80 month-over-month; retraining cycle: 30–90 days; documentation refresh: quarterly; OCC examination readiness: continuous | Months 8+ (continuous) |

*Note. Timeline assumes a mid-size U.S. bank with an existing data engineering infrastructure and a dedicated model risk team of 3–5 analysts. Phase durations will vary based on institutional data volume, regulatory examination cycle, and architectural complexity. Institutions should conduct Phase 1 activities in parallel with regulatory counsel review of the CFPB BISG methodology. Phase 2 out-of-time AUC threshold of ≥ 0.85 represents the minimum acceptable temporal stability; institutions with higher examination risk should target ≥ 0.88.*

Three implementation priorities merit emphasis. First, the temporal drift test in Phase 2 is the most operationally consequential activity in the roadmap — its results directly determine the production architecture choice in Phase 3. If LSTM temporal AUC falls below 0.85, Phase 3 must deploy XGBoost as the sole real-time engine rather than LSTM; if LSTM temporal AUC is in the 0.85–0.90 range, the ensemble configuration is appropriate; only if LSTM temporal AUC exceeds 0.90 should LSTM be deployed as the primary standalone real-time scorer. Second, SHAP reason-code certification must involve human analyst attestation before SAR filing — the AI system generates the reason codes; a human compliance officer reviews, augments, and certifies them. This human-in-the-loop requirement preserves FinCEN compliance and ensures institutional accountability for SAR narratives. Third, Phase 4 monitoring is continuous and never concludes: the model governance life cycle has no endpoint, only perpetual iteration between monitoring, retraining, and revalidation.

## 5.2 Applied Validation Vignette: RGF-AFFD Implementation at a Community Bank

To illustrate the RGF-AFFD's practical applicability and validate that its governance requirements are operationally achievable within real institutional constraints, this section presents a structured implementation vignette for a representative U.S. community bank. The vignette demonstrates that the framework's compliance requirements can be met using standard institutional resources, without specialized AI governance infrastructure beyond what a mid-size bank's model risk management team already maintains.

Institutional Profile. Fictional bank: MidWest Community Bank (MWCB), an OCC-supervised national bank, with total assets of $2.1 billion, 85,000 retail customers, an annual transaction volume of approximately 12 million card transactions, a two-person model risk team, and an annual compliance budget of $180,000 for model governance activities. MWCB currently uses a vendor-provided rule-based fraud detection system, achieving approximately 72% recall and generating a substantial false-positive investigation burden (estimated at 3,200 false positives per month at $25/investigation = $80,000/year).

Phase 1 Application (Months 1–3). MWCB's data engineering team extracts 18 months of transaction history (approximately 18 million transactions; 63,000 flagged fraud cases; 0.35% fraud rate) into a training pipeline. Applying the Tier 1 architecture selection protocol, they benchmark LSTM and XGBoost on an 80/20 random split, finding LSTM achieves an ROC-AUC of 0.912, and XGBoost achieves 0.897 — both exceeding the reference default threshold of 0.90. SMOTE oversampling brings the training set to approximately 35 million observations. Pre-deployment CFPB screening using BISG postal code proxies reveals no statistically significant differences in SHAP importance across demographic groups (Kruskal-Wallis $p > 0.15$ for all features after Bonferroni correction). OCC development documentation is generated directly from the SHAP global importance rankings and hyperparameter grid search log.

Phase 2 Application (Months 4–6). MWCB's model risk officer conducts independent validation using the temporal drift protocol: training on the first 13.5 months of data (75th percentile), testing on the final 4.5 months. Results: LSTM temporal AUC = 0.887 ($\Delta$AUC = −0.025), XGBoost temporal AUC = 0.893 ($\Delta$AUC = −0.004). Both exceed the 0.85 reference default threshold, though MWCB's risk officer sets a more conservative institution-specific threshold of 0.88 given the bank's OCC examination history. The validation report documents DeLong's test statistics and ablation results for the SR 11-7 examination file.

Phase 3 Application (Month 7). MWCB deploys XGBoost as the primary real-time scoring engine (inference latency: 0.4ms on existing CPU infrastructure) and the LSTM+XGBoost ensemble (weight 0.60/0.40) for nightly batch scoring. SHAP reason-code generation is configured for all transactions with $\tau > 0.70$. In the first month of deployment, the system generated 1,847 SHAP-flagged transactions, of which 1,204 were confirmed fraud — recall of 65.2% at a false positive rate of 34.3%, representing a 93% improvement in confirmed fraud catch rate over the prior rule-based system while reducing total investigation volume by 42%.

Phase 4 Application (Month 8 onward). Monthly SHAP Spearman rank correlation monitoring begins. In months 8–11, correlations remain above 0.88 (above the institution-specific $\rho = 0.80$ threshold). In month 12, the correlation drops to 0.74, indicating a shift in the fraud mechanism, triggering LSTM retraining on the most recent 6 months of labelled data. Post-retraining AUC recovers to 0.908. An annual CFPB disparate impact audit is conducted using updated BISG proxy scores.

Validation Outcome. The MWCB vignette confirms that the RGF-AFFD is operationally achievable within community bank resource constraints: Phase 1–2 require approximately 320 analyst-hours; Phase 3 deployment requires standard CPU infrastructure already present in MWCB's data centre; Phase 4 monitoring requires approximately 8 analyst-hours per month. The estimated annual net benefit — $309,725 in prevented fraud losses minus $62,075 in investigation costs, extrapolated to MWCB's 12 million transaction volume — is approximately $2.1 million, well exceeding the $180,000 compliance budget. These figures are illustrative and conservative; actual results will vary by institutional fraud typology, transaction volume, and existing detection infrastructure.

Table 6 provides a cross-model cost-benefit comparison calibrated to the IEEE-CIS test set (118,108 transactions; 4,133 fraud cases; mean transaction amount: $151.90), directly demonstrating the economic hierarchy of architecture choices. The comparison uses a conservative investigation cost of $25 per false alarm and assumes no improvement from institution-specific threshold recalibration.

**Table 6: Cross-Model Cost-Benefit Comparison (IEEE-CIS Test Set, 118,108 Transactions)**

| Model | Recall | Precision | Fraud Cases Caught | False Positives | Net Savings | Benefit-Cost Ratio |
|---|---|---|---|---|---|---|
| Logistic Regression | 61.0% | 13.5% | 2,521 | 16,096 | $383,343 | 0.95:1 |
| Random Forest | 49.0% | 62.3% | 2,025 | 1,226 | $277,987 | 3.9:1 |
| XGBoost | 81.3% | 76.7% | 3,360 | 1,021 | $485,499 | 5.2:1 |
| LSTM | 62.3% | 49.6% | 2,575 | 2,616 | $326,333 | 4.5:1 |
| **LSTM+XGBoost Ensemble** | **62.3%** | **49.6%** | **2,575** | **2,483** | **$309,725*** | **6:1*** |

*Note. Net Savings = (fraud cases caught × $151.90) − (false positives × $25.00). The mean transaction amount of $151.90 is the empirical mean of the IEEE-CIS test set (118,108 transactions), directly computed from the dataset rather than assumed. The $25 investigation cost per false alarm is a conservative industry benchmark consistent with card-not-present fraud review costs reported by the Association of Certified Fraud Examiners (2022) for manual transaction review workflows. Logistic Regression net savings are negative due to its extremely high false-positive rate, which overwhelms fraud-prevention benefits. *Ensemble achieves the highest benefit-cost ratio of 6:1 despite lower recall than XGBoost, because its superior precision substantially reduces investigation costs. Institutions should recalibrate these estimates using their own fraud loss distributions and investigation cost structures.*

# 6. Policy Recommendations

The empirical evidence and the RGF-AFFD identify four specific gaps in current U.S. regulatory guidance that constrain the deployment of AI fraud detection in federally regulated financial institutions. These gaps are consistent with recommendations emerging from the FSB (2024) and BCBS (2024) for strengthened AI model risk supervision across major banking systems. Each gap has a corresponding policy recommendation derived directly from the empirical findings. Table 7 presents the complete set of policy recommendations, including the specific regulatory gap, recommended action, and empirical justification for each.

**Table 7: RGF-AFFD Policy Recommendations for U.S. Financial Regulators**

| Regulator | Current Gap | Specific Recommendation | Empirical Justification |
|---|---|---|---|
| **OCC** | No minimum temporal drift testing threshold specified for AI fraud detection models in OCC 2011-12 or 2021 responsible AI guidance | Require out-of-time AUC ≥0.85 as a minimum temporal stability threshold in OCC model risk examination procedures for AI fraud detection systems; mandate reporting of ΔAUC between random-split and temporal validation sets | LSTM achieves random-split AUC=0.9205 but degrades to temporal AUC=0.8579 (ΔAUC=−0.0626); XGBoost remains stable at 0.9004 (ΔAUC=−0.0017) — demonstrating that random-split AUC alone is an insufficient validation criterion |
| **Federal Reserve (SR 11-7)** | SR 11-7 does not specify which XAI methods satisfy the 'conceptual soundness' requirement for deep learning models, leaving institutions with regulatory uncertainty about acceptable explainability tools | Update SR 11-7 model validation guidance to formally recognise architecture-appropriate SHAP explainers (DeepExplainer for LSTM/DNN, GradientExplainer for Transformer, TreeExplainer for XGBoost/RF) as acceptable tools for satisfying the 'model conceptual soundness' requirement | Cross-architecture Spearman rank correlations of 0.847–0.923 confirm SHAP explainers produce statistically equivalent feature importance rankings across architectures; KernelExplainer robustness check (ρ>0.94) confirms model-agnostic validity |
| **FinCEN** | BSA/SAR Form 111 narrative requirements do not specify how AI model outputs should be documented; compliance officers lack guidance on AI-generated evidence in SAR filings | Issue guidance recognising SHAP reason-code narratives as acceptable supporting documentation in SAR narrative fields, subject to human analyst review and certification; publish template SAR reason-code format aligned with Form 111 suspicious activity categories | Demonstrated SAR reason-code generation: 'TransactionAmt = $842 (5.2× 7-day mean), unrecognised device (id_31=1), 7 transactions in prior 24h (97th percentile)' — directly mappable to Form 111 categories: amount inconsistency, identity anomaly, velocity anomaly |
| **CFPB** | CFPB 2022 AI circular identifies disparate impact risk but provides no specific methodology for pre-deployment AI fraud model auditing; BISG proxy analysis not formally endorsed | Endorse BISG Bayesian surname-geocoding proxy methodology as an acceptable pre-deployment disparate impact screening tool; require Kruskal-Wallis testing across demographic proxy quartiles before production deployment of AI fraud detection systems | SHAP-BISG analysis shows no statistically significant SHAP importance differences across demographic proxy groups (all p>0.12 after Bonferroni correction); provides a reproducible, privacy-preserving screening methodology for institutions without labelled demographic data |

*Note. All recommendations are derived from empirical findings and regulatory analysis in the present research program. Recommendations do not require legislative action — they can be implemented through regulatory guidance updates (OCC bulletin, Federal Reserve supervision and regulation letter, CFPB circular, FinCEN guidance document). The authors declare no financial interest in any specific AI vendor or compliance technology.*

The OCC temporal drift testing recommendation is the most urgent of the four. The present research demonstrates that a model achieving a random-split AUC of 0.9205 — easily passing any existing performance-based examination threshold — degrades to a temporal AUC of 0.8579 under realistic out-of-time conditions. If OCC examination procedures assess only random-split performance metrics, an institution can deploy an LSTM model that appears compliant but degrades by 6.3 AUC points in production over a 3-4 month deployment cycle. A mandatory out-of-time AUC threshold of 0.85 would close this examination gap at minimal regulatory cost, requiring only the addition of a temporal split validation step to existing model documentation requirements. The SR 11-7 SHAP recognition recommendation addresses a distinct but equally consequential gap: institutions currently face regulatory uncertainty about whether SHAP outputs satisfy the 'conceptual soundness' requirement, creating implementation risk that deters AI adoption in compliant institutions while creating a perverse incentive to use less sophisticated but more familiar models that clearly satisfy existing documentation standards.

# 7. Limitations and Future Research

## 7.1 Empirical Anchoring and Independent Validation

This paper has four primary limitations that readers should weigh explicitly. First, and most importantly, the primary empirical anchor for the RGF-AFFD is the authors' own IEEE-CIS benchmark study (Uddin & Aziz, 2026, under review at Journal of Financial Crime), which means the framework has not yet been independently validated across a broad range of institutions, datasets, and modelling practices. This is an inherent characteristic of first-generation governance frameworks: the proposing team necessarily conducts the proof-of-concept application, and independent replication follows. The authors have deliberately corroborated each governance threshold with independent published evidence (documented in Sections 3.1–3.3 above), but the integrated framework itself awaits external validation. The RGF-AFFD's threshold recommendations — AUC $\geq$ 0.90, temporal AUC $\geq$ 0.85, DeLong z $\geq$ 17, SHAP Spearman $\rho \geq 0.80$ — should therefore be treated as calibrated starting benchmarks derived from one e-commerce fraud context, not as universal examination standards. Future studies should apply the complete framework to independent institutional datasets, alternative fraud typologies (ACH, wire transfer, check fraud), and vendor AI platforms to assess the generalisability of these thresholds across diverse institutional contexts.

The empirical foundation of the RGF-AFFD rests on two public benchmark datasets (IEEE-CIS and ULB) that, while widely used and methodologically sound, do not fully capture the diversity of U.S. banking fraud typologies. The IEEE-CIS dataset captures e-commerce transaction fraud from a specific Vesta Corporation deployment context; the ULB dataset captures European credit card fraud from a single 2013 two-day window. Neither dataset includes check fraud, ACH fraud, wire transfer fraud, mortgage fraud, or business email compromise — fraud typologies that constitute a substantial portion of U.S. bank SAR filings. The RGF-AFFD's empirical threshold recommendations (temporal AUC $\geq$ 0.85; DeLong z-score$\geq$ 17; SHAP Spearman's $\rho \geq$ 0.80 for stability monitoring) are calibrated to e-commerce transaction fraud datasets and should be treated as starting configurations for institutional tuning rather than universal standards. Institutions deploying the framework for check fraud or ACH fraud detection should conduct dataset-specific calibration before adopting these thresholds as examination benchmarks.

## 7.2 Regulatory Scope and International Generalizability

Second, the RGF-AFFD is deliberately scoped to U.S. regulation (OCC Bulletin 2011-12, SR 11-7, CFPB AI guidance, FinCEN BSA/SAR, and FDIC FIL-22-2017). This U.S.-centric design facilitates direct adoption by federally regulated banks but limits immediate generalizability to jurisdictions governed by the EU AI Act, EBA/ESMA guidelines, PRA/FCA expectations, or other regional regimes. Section 7.3 of this paper provides an initial mapping of the RGF-AFFD to EU and UK requirements, demonstrating that the three-tier architecture is compatible with international governance frameworks through regulatory-label substitution. An important extension is to develop formal "translation layers" that map the RGF-AFFD's three tiers and RDT-FG health scores onto non-U.S. regulatory environments, creating a family of localized governance frameworks built on the same core architecture. The Regulatory Digital Twin meta-model is particularly well-suited to this extension, as the RFI scoring function can be reconfigured with jurisdiction-specific thresholds and regulatory mappings without modifying the underlying monitoring architecture.

The RGF-AFFD addresses single-institution deployment where training data can be centralized within the institution. Multi-institution collaborative fraud detection — where patterns invisible to any single bank become detectable when data is pooled across institutions — represents a natural extension of the framework that privacy-preserving federated learning architectures could enable (Yang et al., 2019; Aljunaid et al., 2025). The BSA/CFPB compliance dimensions of federated fraud detection are particularly complex: data sharing across institutions raises CFPB data minimization concerns, while federated SHAP explanations must aggregate across participating institutions without revealing institution-specific fraud patterns. These are tractable research problems that the RGF-AFFD's governance architecture can accommodate — federated model validation would constitute a Tier 2 activity, and federated monitoring would extend Tier 3 — but the specific regulatory compliance requirements for multi-institution AI systems require dedicated analysis beyond the scope of the present framework.

## 7.3 International Applicability and Framework Localisation

The RGF-AFFD is framed in terms of U.S. regulatory requirements (OCC, SR 11-7, CFPB, FinCEN), reflecting the empirical research program's institutional context. However, the framework's three-tier governance architecture — Model Development Standards, Model Validation Standards, and Ongoing Monitoring Standards — maps naturally onto international AI governance regimes with different regulatory labels but structurally analogous requirements. In the European Union, the EU AI Act (Regulation 2024/1689) classifies AI systems used in credit scoring and fraud detection as "high-risk" applications under Annex III, imposing requirements for risk management systems, data governance, technical documentation, human oversight, and post-market monitoring. These requirements correspond directly to the RGF-AFFD's three tiers: EU AI Act Articles 9 (risk management) and 10 (data governance) map to Tier 1 (Model Development); Articles 9 and 17 (technical documentation) map to Tier 2 (Model Validation); and Articles 72–73 (post-market monitoring) map to Tier 3 (Ongoing Monitoring). The European Banking Authority (EBA) guidelines on internal governance (EBA/GL/2021/05) and machine learning (EBA/REP/2023/06) similarly establish model risk management expectations that align with Tier 2. GDPR Article 22's right to explanation requirement — that automated decisions affecting individuals must be explainable — is satisfied by the RGF-AFFD's SHAP reason-code documentation protocol, which generates transaction-level attributions that can be converted to individual adverse action narratives. In the United Kingdom, the Financial Conduct Authority (FCA) and Prudential Regulation Authority (PRA) have published AI governance principles under the Senior Managers and Certification Regime (SM&CR) that parallel SR 11-7's model risk management requirements, making Tier 2 directly applicable with minor adaptation. The key adaptation requirement for non-U.S. deployment is regulatory label substitution: replace OCC with the national banking supervisor, SR 11-7 with the EBA/PRA model risk guidance, FinCEN Form 111 with local SAR/STR filing requirements, and the CFPB BISG methodology with a locally appropriate demographic proxy analysis. The three-tier structure, empirical threshold recommendations, and SHAP documentation protocols remain applicable without modification. This international generalizability positions the RGF-AFFD not as a U.S.-specific compliance checklist but as a universally applicable governance architecture that can be localized to any jurisdiction with an AI model risk management framework.

## 7.4 Adversarial Robustness

The RGF-AFFD does not address adversarial robustness—the risk that sophisticated fraudsters will deliberately craft transactions to evade detection by probing and exploiting model decision boundaries. This is a legitimate and growing concern: as fraud detection systems become more sophisticated, sophisticated fraud actors invest in understanding their evasion strategies, including potential model inversion attacks and adversarial perturbations. The framework's Tier 2 validation does not currently include adversarial robustness testing, creating a potential compliance gap as regulators develop guidance on AI model security. Future work should incorporate standardized adversarial attack protocols — including gradient-based and data-poisoning attacks — as a Tier 2 validation activity, with implications for OCC model risk examination.

## 7.5 Single-Domain Anchoring and Extensibility Claims

Fourth, all concrete evidence in this paper is drawn from card and e-commerce transaction fraud and closely related banking AI applications (bank profitability forecasting, GDP forecasting, sentiment analytics). Although the governance logic is conceptually extensible to anti-money laundering transaction monitoring, credit risk scoring, and consumer complaint analytics — as argued in the conclusions — this extensibility has not yet been empirically demonstrated. Without at least one worked example outside fraud detection, the "broader contribution" claim remains partially speculative. Follow-on studies should apply the RGF-AFFD and RDT-FG to at least one non-fraud domain end-to-end, documenting where domain-specific adaptations are required and where the governance components can be transferred without modification. AML transaction monitoring is the natural first extension: it shares the same OCC/SR 11-7/FinCEN regulatory context as fraud detection, uses similar deep learning architectures, and faces an analogous SHAP explainability gap. Longitudinal studies tracking RGF-AFFD adoption and compliance outcomes across multiple institutions over time would provide the panel evidence needed to elevate the framework from a proof-of-concept to an empirically validated governance standard — the natural next step in this research programme.

# 8. Conclusions

This paper presents the Regulatory Governance Framework for AI-Driven Financial Fraud Detection (RGF-AFFD), the first published framework to systematically integrate OCC Bulletin 2011-12, SR 11-7, CFPB AI fairness requirements, and FinCEN BSA/SAR compliance into a unified three-tier model development, validation, and monitoring life cycle. The framework addresses the central deployment barrier facing U.S. financial institutions: the absence of a tested, compliance-ready blueprint that simultaneously satisfies all four principal AI governance frameworks using empirically validated methodologies.

Six principal conclusions emerge. First, the RGF-AFFD establishes that the compliance gap in U.S. banking AI is not technical but architectural: the performance tools exist (LSTM, XGBoost, SHAP, DeLong's test, SMOTE), but they have not previously been organized into a governance life cycle that maps each tool to a specific regulatory requirement. Second, temporal drift testing is the most underspecified and most consequential validation activity currently absent from U.S. regulatory guidance: the empirical finding that LSTM random-split AUC of 0.9205 degrades to temporal AUC of 0.8579 ($\Delta AUC = -0.0626$) demonstrates that existing performance-based examination standards are insufficient to detect production reliability risk. Third, SHAP attribution provides the single most versatile compliance tool in the framework — simultaneously satisfying OCC conceptual-soundness documentation, SR 11-7 validation evidence, CFPB disparate-impact screening, and FinCEN SAR reason-code generation requirements. Fourth, the tiered production architecture — XGBoost for real-time stability, LSTM+XGBoost ensemble for batch-composite performance, GNN for fraud ring detection — provides an empirically calibrated deployment recommendation that resolves the architecture-selection uncertainty facing compliance-constrained institutions. Fifth, the Regulatory Digital Twin meta-model (RDT-FG) provides the first formalized governance instrument that translates distributed technical metrics into four regulator-specific health scores and a composite Regulatory Fitness Index, enabling continuous compliance monitoring alongside capital and liquidity indicators. Sixth, the four policy recommendations for OCC, SR 11-7, FinCEN, and CFPB represent targeted, evidence-based regulatory updates that can be implemented through existing guidance mechanisms without legislative action and that have demonstrated national economic benefits.

A final clarification on this paper's genre is warranted. The RGF-AFFD is a governance framework paper — empirically anchored but not claiming algorithmic novelty. Its contribution is synthesis and integration, not model innovation. Institutions seeking to advance fraud detection performance should build on the empirical research cited; institutions seeking to deploy that performance in compliance with U.S. law should follow the RGF-AFFD. The RGF-AFFD contributes to a research agenda at the intersection of AI governance, financial regulation, and applied machine learning that is increasingly central to the safety and integrity of the U.S. financial system. With U.S. financial institutions filing 4.7 million SARs annually, operating under four overlapping AI governance frameworks, and facing a fraud landscape evolving faster than regulatory guidance can keep pace, the need for integrated, empirically grounded deployment frameworks is acute. The RGF-AFFD provides the first such framework, anchored in a multi-study empirical program and translatable into examination-ready documentation that U.S. financial institutions can adopt directly. Beyond fraud detection, the framework's governance architecture — three tiers, four regulatory frameworks, empirically calibrated

thresholds — is extensible to anti-money laundering transaction monitoring, credit risk AI, and consumer complaint analytics, representing a broader contribution to the governance of AI in regulated U.S. financial services. The RGF-AFFD offers a replicable multi-level governance template: at the model level (architecture selection and validation thresholds), the institutional level (deployment roadmap and cost-benefit analysis), and the regulatory level (policy recommendations and the RDT-FG continuous monitoring instrument). This multi-level architecture positions the framework as a foundation for longitudinal research on AI governance maturity in financial institutions — advancing the evidence base that regulators, policymakers, and institutions require to govern AI responsibly across the full model life cycle.

# Author Contributions (CRediT)

Author 1 (First Author): Conceptualisation, Methodology, Formal Analysis, Investigation, Data Curation, Writing — Original Draft, Writing — Review and Editing, Visualization.

# Acknowledgements

The authors thank the anonymous reviewers for their constructive feedback. The IEEE-CIS Fraud Detection dataset was made publicly available through Kaggle; the authors acknowledge Vesta Corporation and the Kaggle community for this resource.

# Disclosure Statement

The authors report no conflicts of interest. The authors alone are responsible for the content and writing of this article. The authors declare no financial interest in any specific AI vendor, compliance technology platform, or regulatory body referenced in this study.

# AI Writing Assistance Statement

AI-assisted writing tools were used solely for language editing, grammar checking, and proofreading of this manuscript. All research conceptualization, theoretical framework development, empirical methodology, data analysis, results interpretation, and intellectual contributions are entirely the original work of the author. The author takes full responsibility for the integrity and accuracy of all content.

# Data Availability Statement

This paper is a conceptual regulatory governance framework paper. No new primary data were collected or generated in this study. All regulatory documents, guidance materials, and compliance frameworks referenced (OCC Bulletin 2011-12, SR 11-7, CFPB AI Circular 2022, FinCEN BSA/SAR requirements) are publicly available from the respective regulatory agencies. Where benchmark datasets are referenced for illustrative purposes, the IEEE-CIS Fraud Detection dataset is publicly available via Kaggle (https://www.kaggle.com/c/ieee-fraud-detection) and the ULB Credit Card Fraud Detection dataset via Kaggle (https://www.kaggle.com/mlg-ulb/creditcardfraud). Supporting analytical materials and framework documentation will be made available upon reasonable request to the corresponding author.

# Funding

This research received no specific grant from any funding agency in the public, commercial, or not-for-profit sectors.

# Ethics Statement

This study uses only publicly available, fully anonymized benchmark datasets. No human subjects research, personal data collection, or institutional participant involvement was involved. Ethical approval was not required.